\documentclass[reprint,floatfix,superscriptaddress,nofootinbib,amsmath,amssymb,aps,longbibliography]{revtex4-2}

\usepackage[utf8]{inputenc}
\usepackage[T1]{fontenc}
\usepackage{microtype}
\usepackage{booktabs}
\usepackage{graphicx}
\graphicspath{{figures/}{code/results/}}
\usepackage{xcolor}
\usepackage{listings}
\usepackage{tikz}
\usetikzlibrary{arrows.meta, positioning, calc}
\usepackage[colorlinks=true,linkcolor=blue!50!black,citecolor=blue!50!black,urlcolor=blue!50!black]{hyperref}
\usepackage{xurl} % let long URLs in the bibliography break anywhere

\definecolor{figblue}{HTML}{2A78D6}
\definecolor{figink}{HTML}{0B0B0B}
\definecolor{figmuted}{HTML}{52514E}
\definecolor{figgrid}{HTML}{D9D8D4}

\lstdefinestyle{spkpython}{
  language=Python,
  basicstyle=\ttfamily\footnotesize,
  keywordstyle=\color{blue!50!black}\bfseries,
  commentstyle=\color{green!35!black}\itshape,
  stringstyle=\color{red!45!black},
  showstringspaces=false,
  breaklines=true,
  frame=single,
  framesep=3pt,
  framerule=0.4pt,
  rulecolor=\color{black!30},
  numbers=none,
  tabsize=4,
  captionpos=b,
  belowcaptionskip=4pt,
  aboveskip=6pt,
  belowskip=6pt,
}
\newtheorem{definition}{Definition}
\begin{document}
\raggedbottom

\title{Quantum-inspired orthogonal adapters for certifiably robust medical image classification}

\author{Olga Tsurkan}
\affiliation{EigenOrbit Sàrl, Luxembourg}

\author{Laia Domingo}
\affiliation{EigenOrbit Sàrl, Luxembourg}

\author{Victoria Orlova}
\affiliation{Dermatology Clinic, Hospital Maggiore of Trieste, University of Trieste, Trieste, Italy}

\author{Asel Sagingalieva}
\affiliation{EigenOrbit Sàrl, Luxembourg}

\author{Iris Zalaudek}
\affiliation{Dermatology Clinic, Hospital Maggiore of Trieste, University of Trieste, Trieste, Italy}

\author{Raffaele Dante Caposiena Caro}
\email{dcaposiena@hotmail.com}
\affiliation{Dermatology Clinic, Hospital Maggiore of Trieste, University of Trieste, Trieste, Italy}

\begin{abstract}

Medical image classifiers require not only high accuracy but also reliable predictions under small input perturbations. Certified robustness provides mathematical guarantees of prediction stability, but preserving these guarantees when adapting pretrained models to new tasks remains challenging. We propose a fine-tuning approach based on quantum-inspired orthogonal butterfly layers that preserves the Lipschitz bounds of pretrained certified networks. Inspired by quantum circuits of reconfigurable beam splitters, these layers preserve orthogonality by construction using only $\mathcal{O}(n\log n)$ trainable parameters. Their structured design enables efficient classical implementation and offers a path toward future quantum implementations with logarithmic circuit depth under suitable connectivity assumptions. Controlled experiments on MNIST and MedMNIST demonstrate improved certified robustness compared with conventional alternatives and, in the DermaMNIST transfer, greater resistance to the tested attacks than low-rank adaptation (LoRA). We then apply it to trichoscopic image classification for alopecia diagnosis and alopecia areata activity assessment. By adapting a certified ImageNet-pretrained network, butterfly fine-tuning achieves approximately twice the certified accuracy of conventional fine-tuning and LoRA, with comparable classification accuracy among the adaptation methods evaluated on the same backbone and data split. Larger robustness guarantees, including certificates covering certain image-processing transformations, can be obtained at the cost of reduced accuracy. In these experiments, butterfly adaptation achieved higher certified accuracy than LoRA and unrestricted fine-tuning, while the methods showed similar empirical resistance to the tested adversarial attacks. These results highlight the potential of quantum-inspired butterfly layers for parameter-efficient adaptation of medical image classifiers while preserving the mathematical guarantees underlying certified robustness.

\end{abstract}

\maketitle

\section{Introduction}
\label{sec:introduction}

Deep learning has achieved remarkable performance in medical image analysis in many different fields, including dermatology. For hair-loss disorders, deep neural networks have been used to measure the extent of hair loss \cite{lee2020clinically,gudobba2023automating} and hair density \cite{kim2022evaluation}, and to analyse scalp images \cite{kim2023deep}. They have also been applied to trichoscopic images to classify androgenetic alopecia \cite{gao2022deep}, diagnose hair-loss disorders, and assess alopecia areata activity \cite{caro2026trichoscopy}.
Their adoption in clinical practice, however, raises an important question: \emph{are their predictions robust to small image perturbations?} A model may achieve high test accuracy while remaining sensitive to small, even imperceptible, changes that alter its predicted diagnosis \cite{finlayson2019adversarial}. Indeed, stress tests of dermatology classifiers have shown that even simple changes, such as rotating an image or photographing the same area again, can affect their predictions \cite{young2021stress,maron2021robustness}. This sensitivity raises concerns about their safety in clinical use \cite{kelly2019key}, and regulators now explicitly require or recommend robustness \cite{euaiact2024,fda2025aienabled}. Test accuracy alone cannot establish whether a particular prediction will remain stable when the input changes.

Certified robustness addresses this concern by providing a mathematical guarantee: for each image, any input perturbation within a certified radius, measured under a chosen norm, leaves the model's prediction unchanged \cite{wong2018provable,cohen2019certified,li2023sok}. One way to obtain such a certificate is to bound the Lipschitz constant of the network, which, combined with the classification margin, allows the certified radius to be computed for each prediction using a single forward pass \cite{tsuzuku2018lipschitz,anil2019sorting}. While this does not guarantee that the predicted diagnosis is medically correct, it establishes that the prediction is stable. In practice, obtaining useful Lipschitz bounds often requires architectures specifically designed to control them. For image classification, networks such as BRONet \cite{lai2025bronet} have demonstrated meaningful certified robustness on ImageNet. However, applying these networks to specialized medical domains presents an additional challenge. Medical datasets are often too small to train such models from scratch, making it necessary to adapt pretrained networks through fine-tuning. Conventional fine-tuning, however, can alter their Lipschitz bounds and compromise the original robustness guarantees.

Orthogonal transformations offer a way to preserve Lipschitz bounds during fine-tuning. Because they preserve Euclidean norms, multiplying a pretrained layer's weight matrix by an orthogonal matrix leaves its Lipschitz bound unchanged \cite{cisse2017parseval,anil2019sorting,li2019preventing,prach20241lipschitz}. This allows the network to adapt to new tasks while retaining the Lipschitz constraints required for certified robustness. Quantum-inspired circuits provide orthogonal transformations by construction. In particular, circuits built from Reconfigurable Beam Splitter (RBS) gates combine simple rotations between pairs of coordinates while preserving orthogonality for any choice of parameters \cite{kerenidis2021orthogonal,landman2022quantum}. Arranging these gates in a butterfly structure makes the representation particularly efficient: a layer of width $n$ requires only $\tfrac{n}{2}\log_2 n$ rotations, organized into $\log_2 n$ parallel stages \cite{cherrat2024quantum}. This substantially reduces the number of trainable parameters compared with a general orthogonal matrix. These transformations can be computed efficiently on classical hardware, as in all our experiments \cite{landman2022quantum}. Their direct correspondence to quantum circuits also opens the possibility of future quantum implementations, although any practical advantage would depend on data encoding, hardware connectivity, and measurement requirements \cite{kerenidis2021orthogonal,landman2022quantum}.

Structured orthogonal transformations have also been explored for a different purpose: parameter-efficient fine-tuning of pretrained models. Examples include orthogonal fine-tuning \cite{qiu2023oft}, its butterfly \cite{liu2024boft} and quasi-orthogonal Givens \cite{ma2024qgoft} variants, and quantum-inspired compound adapters \cite{raj2025quic}, which offer alternatives to widely used methods such as Low Rank Adaptation (LoRA) \cite{hu2022lora}. However, these approaches focus on efficient adaptation rather than certified robustness. To our knowledge, orthogonal adapters have not previously been used to preserve the Lipschitz bounds of certified networks during fine-tuning.\\

\begin{figure*}[t]
    \centering
    \includegraphics[width=\textwidth]{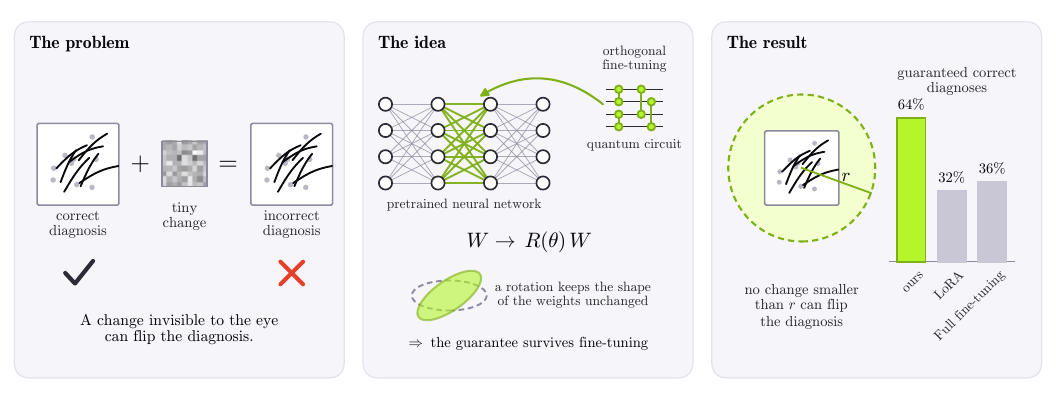}
    \caption{Overview of the work.
    The problem: a change of a trichoscopic image that is invisible to the eye can change the diagnosis of a neural network.
    The idea: the weights $W$ of a pretrained certified network are fine-tuned by multiplying them with an orthogonal matrix $R(\theta)$ implemented as a quantum-inspired circuit; the rotation keeps the singular values of the weights, so the robustness guarantee survives fine-tuning.
    The result: no change of the image smaller than the certified radius $r$ can change the diagnosis, and with orthogonal fine-tuning this guarantee holds for about twice as many correct diagnoses as with LoRA or full fine-tuning (diagnosis task, $\varepsilon=36/255$).}
    \label{fig:graphical_abstract}
\end{figure*}

In this work, we investigate quantum-inspired butterfly layers for building and adapting certifiably robust neural networks, with trichoscopy as our main clinical application, as illustrated in Fig.~\ref{fig:graphical_abstract}. We first construct and train butterfly-based networks from scratch on MNIST and MedMNIST, showing that their orthogonality maintains useful robustness guarantees even as network depth increases. In contrast, spectral normalisation can attenuate the signal, while unconstrained layers cannot provide comparable guarantees. In the controlled DermaMNIST transfer, butterfly-adapted models were also more robust to the tested attacks than LoRA models.
Our main contribution is a parameter-efficient fine-tuning approach that preserves the Lipschitz bounds of pretrained certified networks. We demonstrate this by adapting an ImageNet-pretrained BRONet to trichoscopic image classification for alopecia diagnosis and alopecia areata activity assessment. Butterfly fine-tuning achieves approximately twice the certified accuracy of conventional fine-tuning and LoRA, with comparable classification accuracy among the adaptation methods evaluated on the same backbone and data split. We further show that larger certificates, including radii covering certain image-processing transformations, can be obtained at the cost of reduced accuracy. In these experiments, butterfly adaptation achieved higher certified accuracy than LoRA and unrestricted fine-tuning, while the methods showed similar empirical resistance to the tested adversarial attacks.

The remainder of this paper is organised as follows. Section~\ref{sec:related_work} reviews related work. Section~\ref{sec:methodology} introduces the certification framework, orthogonal butterfly networks, and orthogonal fine-tuning strategies, while Sec.~\ref{sec:experimental_setup} describes the datasets and experimental protocols. Section~\ref{sec:results} presents the results, and Sec.~\ref{sec:conclusions} discusses their clinical implications, the limitations of this work, and the main conclusions. Additional experiments with a non-certified backbone and adversarial attack curves are provided in the Appendices.

% ============================================================
\section{Related Work}
\label{sec:related_work}
% ============================================================

Our work connects four lines of research: deep learning in dermatology and its reliability, certified robustness, quantum orthogonal layers, and orthogonal fine-tuning of pretrained models.
We review them in this order and, for each, state how our work relates to it.

\subsection{Deep Learning in Dermatology}
\label{subsec:rw_dermatology}

Deep learning models have achieved dermatologist-level performance in certain skin lesion classification tasks \cite{esteva2017dermatologist,haenssle2018man}, and their use is expanding across medical imaging \cite{litjens2017survey,zhou2021review,topol2019high}. Trichoscopy, a dermoscopic technique for examining hair and scalp, is widely used to diagnose hair-loss disorders \cite{rudnicka2011trichoscopy,inui2011trichoscopy}, but its application in deep learning remains relatively limited \cite{gao2022deep,caro2026trichoscopy}. In particular, Caposiena Caro \emph{et al.}\ \cite{caro2026trichoscopy} investigated convolutional neural networks for distinguishing different forms of alopecia from healthy scalp and assessing alopecia areata activity. Their study provides the clinical foundation for our work.

Beyond classification accuracy, previous studies have identified important limitations in the reliability of medical image classifiers. Models may rely on confounding features and generalise poorly across hospitals \cite{zech2018variable}, while overall performance metrics can conceal failures in clinically relevant subgroups \cite{oakdenrayner2020hidden}. Similar concerns have been reported in dermatology, where classifiers show reduced performance across diverse skin tones and external datasets \cite{daneshjou2022disparities}. Robustness benchmarks have also revealed sensitivity to image corruptions and perturbations \cite{maron2021benchmark}. However, these evaluations assess reliability empirically, without providing mathematical guarantees for individual predictions. To our knowledge, our work is the first to investigate certified robustness in trichoscopic image classification.

\subsection{Certified Robustness}
\label{subsec:rw_certified}

Deep neural networks can change their predictions in response to small, carefully chosen perturbations of their inputs \cite{szegedy2014intriguing,goodfellow2015explaining}. This sensitivity  is commonly evaluated using adversarial attacks, such as projected gradient descent \cite{madry2018towards}, Carlini--Wagner attacks \cite{carlini2017towards}, and AutoAttack \cite{croce2020reliable,andriushchenko2020square}, with standardized benchmarks enabling systematic comparisons \cite{croce2021robustbench}. However, resistance to known attacks does not guarantee robustness, as stronger or previously unknown attacks may still succeed \cite{athalye2018obfuscated}.

Certified robustness addresses this limitation by mathematically guaranteeing that a prediction remains unchanged within a specified perturbation radius \cite{li2023sok}. Since exact verification is an NP-complete problem, being computationally difficult for large networks \cite{katz2017reluplex}, several approaches have been developed. Bound propagation methods estimate how perturbations affect the outputs of successive layers \cite{wong2018provable,gowal2019scalable,zhang2020towards}, while randomized smoothing provides probabilistic guarantees through repeated evaluations under random noise \cite{lecuyer2019certified,cohen2019certified}. Alternatively, Lipschitz-based methods bound the network's sensitivity to input changes \cite{tsuzuku2018lipschitz,leino2021globally}, allowing a deterministic certificate to be computed from the classification margin in a single forward pass (see Sect. \ref{sec:methodology}). Recent advances have extended this approach to ImageNet-scale classification \cite{hu2023unlocking,lai2025bronet}.
Such networks are built from spectrally normalised layers \cite{miyato2018spectral,gouk2021regularisation} or from orthogonal and other 1-Lipschitz layers \cite{cisse2017parseval,trockman2021orthogonalizing,singla2021skew,xu2022lot,prach2022almost,araujo2023unified,meunier2022dynamical,lai2025bronet}, which differ substantially in memory, speed and certified accuracy \cite{prach20241lipschitz}.
Orthogonal layers keep the norm of every input and gradient, so that the signal does not fade in deep networks \cite{saxe2014exact,xiao2018dynamical,anil2019sorting,li2019preventing}.

In medical imaging, previous studies have demonstrated the vulnerability of classifiers to adversarial perturbations \cite{finlayson2019adversarial,paschali2018generalizability,bortsova2021adversarial,ma2021understanding,hirano2021universal}. Research on certified robustness has also explored randomized smoothing, including applications to segmentation \cite{laousy2023certification} and vision--language classification \cite{hussein2024promptsmooth}. Our work instead focuses on deterministic Lipschitz-based certification and, specifically, on preserving the Lipschitz bounds of pretrained certified networks during fine-tuning for specialized medical classification tasks.

%\subsection{Lipschitz-Constrained and Orthogonal Networks}
\subsection{Quantum Orthogonal Layers}
\label{subsec:rw_quantum}
Quantum machine learning \cite{biamonte2017quantum,cerezo2021variational} has explored circuits of Reconfigurable Beam Splitter (RBS) gates for data encoding and constructing orthogonal neural-network layers \cite{johri2021nearest,kerenidis2021orthogonal,landman2022quantum}. These gates perform rotations between pairs of states, producing orthogonal transformations when operating in the unary (single-excitation) subspace. Different arrangements of these gates, including pyramid, butterfly, and X circuits, offer different trade-offs between the number of parameters and circuit depth \cite{landman2022quantum,cherrat2024quantum}. Landman \emph{et al.}\ \cite{landman2022quantum} demonstrated their use in orthogonal neural networks for medical image classification, while Cherrat \emph{et al.}\ \cite{cherrat2024quantum} extended these constructions to compound matrices for quantum vision transformers. Their trainability and expressivity have also been investigated \cite{monbroussou2025trainability}.

An attractive feature of these architectures is their connection between classical and quantum computing. RBS circuits restricted to the unary subspace can be simulated efficiently on classical hardware \cite{landman2022quantum,monbroussou2025trainability,cerezo2025does}, allowing their properties to be explored without requiring quantum devices. At the same time, their direct correspondence to quantum circuits provides a natural foundation for future quantum implementations and extensions beyond the classically tractable unary setting. In this work, we exploit this connection through a classical implementation that preserves orthogonality by construction. While previous studies have primarily explored classification performance and computational efficiency, we investigate a different application: preserving the Lipschitz bounds required for certified robustness when fine-tuning pretrained neural networks.

\subsection{Orthogonal Fine-Tuning}
\label{subsec:rw_peft}

Parameter-efficient fine-tuning (PEFT) adapts pretrained models by updating only a small number of parameters, reducing the computational and memory costs of training \cite{houlsby2019parameter,ding2023parameter,han2024parameter}. Although widely explored in machine learning, its application to medical imaging remains relatively limited \cite{dutt2024parameter}. Among these methods, LoRA \cite{hu2022lora} introduces low-rank updates to pretrained weights, with extensions such as DoRA \cite{liu2024dora}. Orthogonal fine-tuning instead applies learned orthogonal transformations to pretrained weights \cite{qiu2023oft}. More compact variants have been developed using butterfly factorisations \cite{liu2024boft}, Givens rotations \cite{ma2024qgoft}, and Householder reflections \cite{yuan2024bridging}.

The approach closest to ours is QuIC \cite{raj2025quic}, which constructs fine-tuning adapters from Hamming-weight-preserving quantum circuits and compound matrices. However, QuIC and other orthogonal fine-tuning methods focus primarily on parameter and memory efficiency rather than certified robustness. Our work explores a different application of orthogonal adapters: preserving the robustness certificate of pretrained certified networks during fine-tuning.
% ============================================================
\section{Methodology}
\label{sec:methodology}
% ============================================================

\begin{figure*}[t]
    \centering
    \includegraphics[width=\textwidth]{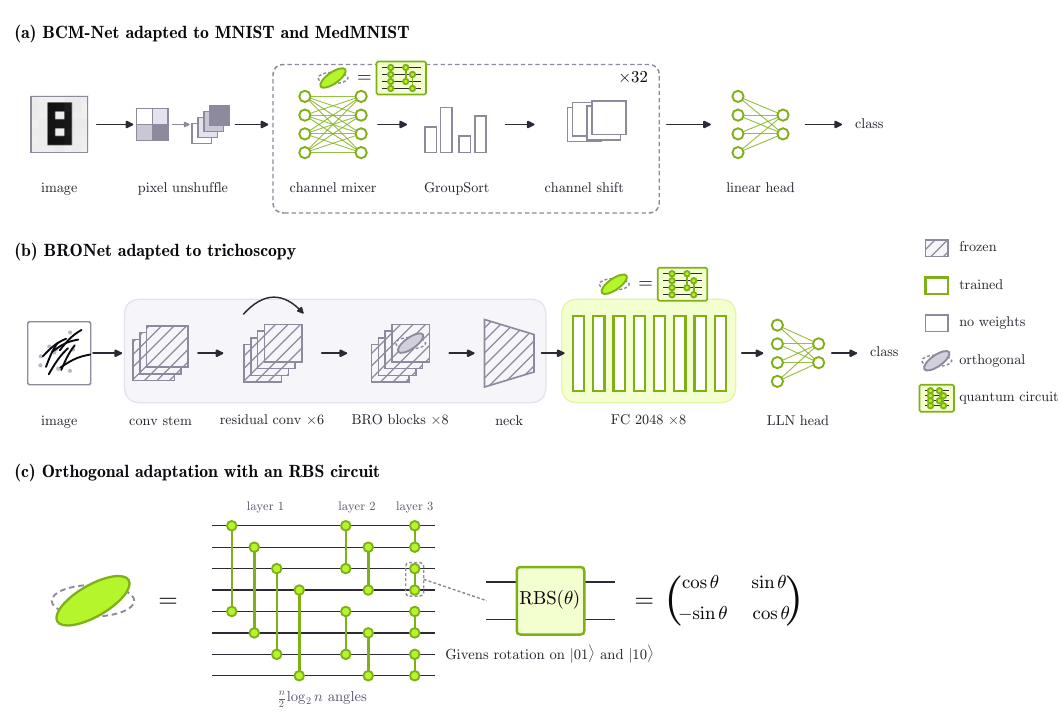}
    \caption{The two certified networks that we adapt and the orthogonal adapter.
    (a)~BCM-Net with 32 blocks, adapted in the controlled transfers on MNIST and MedMNIST: the adaptation strategy acts on all 32 channel mixers, and a new linear head is trained.
    (b)~BRONet adapted to trichoscopy: the convolutional part and the neck are frozen, the strategy acts on the eight fully connected layers, and a new last-layer-normalised (LLN) head is trained.
    Hatched: frozen; blue: trained; light outline: operations without weights.
    The ellipse marks an orthogonal map: grey on the orthogonal BRO blocks, blue where the orthogonal adapter $R(\theta)$ acts.
    (c)~The blue orthogonal maps are implemented as butterfly circuits of reconfigurable beam splitter (RBS) gates, shown for $n=8$ wires: $\log_2 n$ layers and $\tfrac{n}{2}\log_2 n$ trainable angles.
    On the two unary states $|01\rangle$ and $|10\rangle$, each RBS gate acts as a $2\times2$ Givens rotation, so the resulting $n\times n$ matrix is orthogonal for every value of the angles.}
    \label{fig:method}
\end{figure*}

This section explains how a robustness guarantee is obtained for a single prediction, why orthogonal layers keep this guarantee when a network is trained or adapted, and how such layers are built from quantum-inspired circuits.
Figure~\ref{fig:method} summarises the approach.

\subsection{Certified Robustness from a Lipschitz Bound}
\label{subsec:lipschitz_certification}

\begin{definition}[Lipschitz bound]
A function $h$ is $L$-Lipschitz if
\begin{equation}
    \|h(x)-h(x')\|_2 \leq L\,\|x-x'\|_2
    \quad  \forall  x, x'.
    \label{eq:lipschitz}
\end{equation}
The constant $L$ is then called a Lipschitz bound of $h$.
\end{definition}
The constant $L$ is the largest factor by which the function can enlarge a change of its input.
If the input image changes by $\delta$, the features change by at most $L\|\delta\|_2$.

Consider a classifier that computes features $h(x)$ with an $L$-Lipschitz network and then the logits $z_k(x)=w_k^\top h(x)+b_k$ of the classes $k=1,\dots,K$. Let $p=\arg\max_k z_k(x)$ be the predicted class. Then, the corresponding certified radius is defined as follows:
\begin{definition}[Certified radius]
The certified radius of the input $x$ is
\begin{equation}
    r(x)=\min_{j\neq p}
    \frac{z_p(x)-z_j(x)}{L\,\|w_p-w_j\|_2}.
    \label{eq:radius}
\end{equation}
\end{definition}
A perturbation $\delta$ can reduce the difference $z_p-z_j$ by at most $L\|w_p-w_j\|_2\,\|\delta\|_2$, so the prediction cannot change as long as $\|\delta\|_2<r(x)$ \cite{tsuzuku2018lipschitz,singla2022improved}.
In other words, if an image has certified radius $r(x)$, then any change of this image with $\ell_2$ norm smaller than $r(x)$, in any direction, leaves the predicted class unchanged.
The radius is computed for every image from a single forward pass of the network.

The certified radius depends on how confidently the model predicts a class and how sensitive it is to changes in the input. A larger difference between the predicted class and the other classes gives a larger radius, while a larger Lipschitz bound $L$ gives a smaller one. This is why we need to control $L$, both during training and when adapting a model to a new task.

We use two ways to calculate the certified radius. For the networks trained from scratch, we use a simpler, conservative formula:
\begin{equation}
    r(x) \geq \frac{z_p-z_{j^*}}{\sqrt{2}L},
\end{equation}
where $p$ is the predicted class and $j^*$ is the class with the second-highest score. This formula applies when the final classification layer has spectral norm at most one \cite{tsuzuku2018lipschitz}. For pretrained networks, we use the more precise formula in Eq.~\eqref{eq:radius}, which also accounts for the weights of the final classification layer \cite{singla2022improved,lai2025bronet}.\\

\emph{Certified versus empirical robustness.}
The robustness of a classifier is usually measured in one of two ways.
\emph{Empirical} robustness is measured with adversarial attacks, such as projected gradient descent (PGD) \cite{madry2018towards} or its stronger variants \cite{croce2020reliable}, which search for a perturbation of a given norm $\varepsilon$ that changes the prediction.
The robust accuracy under attack is the fraction of test images that are classified correctly and that the attack fails to change.
A failed attack shows only that \emph{this} attack did not find a perturbation; a stronger attack may still find one.
For this reason, many defences that appeared robust under the attacks of their time were later broken by better ones \cite{carlini2017towards,athalye2018obfuscated}. \emph{Certified} robustness is measured with the radius of Eq. \eqref{eq:radius}.
The certified accuracy at $\varepsilon$ is the fraction of test images that are classified correctly and have $r(x)>\varepsilon$.
For these images, no perturbation of norm up to $\varepsilon$ can change the prediction, whatever attack is used.
The true robust accuracy of a model therefore lies between its certified accuracy, a lower bound, and its accuracy under the strongest attack available, an upper bound. \\

The same holds for single images.
Let $d^*(x)$ be the norm of the smallest perturbation that changes the prediction for $x$.
It cannot be computed exactly for deep networks, but it is bounded from both sides,
\begin{equation}
    r(x)\;\le\; d^*(x)\;\le\;\varepsilon^*(x),
    \label{eq:sandwich}
\end{equation}
where $\varepsilon^*(x)$ is the norm of the smallest successful perturbation that an attack finds.
The ratio $\varepsilon^*(x)/r(x)\ge1$ shows how close the certificate is to the robustness that is actually there: a ratio near one means that the certificate describes the model well, and a large ratio means that the bound is loose. An attack that succeeds below $r(x)$ would contradict the certificate; checking that this never happens is a test of the implementation.

\subsection{Orthogonal Layers}
\label{subsec:orthogonal_layers}

Equation~\eqref{eq:radius} shows that a useful certificate requires a small Lipschitz bound.
In a deep network, this bound is built up layer by layer, so the question becomes how to constrain each layer without destroying what the network has learned.
A deep network is a composition $f=f_D\circ\cdots\circ f_1$ of layers.
Lipschitz bounds multiply along a composition,
\begin{equation}
    L_f \leq \prod_{i=1}^{D} L_i,
    \label{eq:composition}
\end{equation}
so a network has a small bound if each of its layers has one.
For a linear layer $x\mapsto Wx$, the bound is the largest singular value $\sigma_{\max}(W)$.

There are two common ways to make a linear layer 1-Lipschitz.
\emph{Spectral normalisation} divides $W$ by $\sigma_{\max}(W)$ \cite{miyato2018spectral}.
This fixes the largest singular value to one but leaves all others free to be smaller, so the layer reduces the norm of its input along some directions.
An \emph{orthogonal} matrix, with $W^\top W=I$, has all singular values equal to one:
\begin{equation}
    \|Wx\|_2=\|x\|_2, \qquad \|W^\top g\|_2=\|g\|_2 .
    \label{eq:norm_preservation}
\end{equation}
It keeps the norm of every input \cite{cisse2017parseval} and of every gradient that flows back through it \cite{anil2019sorting,li2019preventing}.
This difference matters for depth.
If each spectrally normalised layer scales a direction by a factor $\sigma<1$, a stack of $D$ such layers can scale it by $\sigma^D$, and the signal that reaches the classifier attenuates with depth.
Orthogonal layers do not have this problem: orthogonal initialisation preserves signal propagation and allows stable training of very deep networks \cite{saxe2014exact,pennington2017resurrecting,xiao2018dynamical}, and layers that stay orthogonal are used to build deep Lipschitz-bounded networks \cite{anil2019sorting,li2019preventing,trockman2021orthogonalizing,singla2021skew,prach20241lipschitz}.

\subsection{Quantum-Inspired Butterfly Orthogonal Layers}
\label{subsec:butterfly_layers}

Orthogonal layers are thus well suited to certified networks, but they are only useful if they remain orthogonal while they are trained.
Circuits of quantum gates offer a simple way to achieve this, because they are built from rotations.

The reconfigurable beam splitter (RBS) gate rotates the two states $|01\rangle$ and $|10\rangle$ of two qubits by an angle $\theta$ and leaves $|00\rangle$ and $|11\rangle$ unchanged (see Fig.~\ref{fig:method}c) \cite{johri2021nearest,kerenidis2021orthogonal,landman2022quantum}.
Restricted to the $n$ unary states of $n$ qubits, a circuit of RBS gates is a product of Givens rotations, that is, an $n\times n$ orthogonal matrix for \emph{every} value of its angles \cite{landman2022quantum,cherrat2024quantum}, so the layer stays exactly orthogonal throughout training.
We arrange the gates as a butterfly circuit \cite{cherrat2024quantum}, which follows the pattern of the fast Fourier transform (Fig.~\ref{fig:method}c): it has $\tfrac{n}{2}\log_2 n$ angles in $\log_2 n$ layers of parallel rotations, which for $n=2048$ gives $11{,}264$ angles instead of about $4.2$ million weights.
All our experiments run on classical hardware, and we build the matrix from its angles once per forward pass.
What we take from the quantum construction is its \emph{structure}: a cheap parameterisation that is orthogonal by construction and that maps directly to a shallow circuit of RBS gates.

We use butterfly layers in two settings: as the layers of a network trained from scratch, and as adapters of a network that has already been trained.

\subsubsection{Orthogonal Networks Trained from Scratch}
\label{subsubsec:orthogonal_networks}

In the first setting, we want to study orthogonality in isolation.
We therefore use a small network in which the channel-mixing layer is the only linear map that can change the norm of the signal: the image is rearranged into channels by pixel unshuffle, passed through $D$ blocks of a $C\times C$ channel mixer, the GroupSort activation \cite{anil2019sorting} and a fixed channel shift, and classified by a normalised linear layer (see Fig.~\ref{fig:method}a).
All other operations keep norms exactly or are 1-Lipschitz, so the Lipschitz bound of the network is the product of the mixers' bounds. 

In our experiments, we compare three mixers inside the same network: (i)~\texttt{butterfly}, a butterfly circuit of RBS gates ($L=1$ exactly), which we call the Butterfly Channel Mixer network (BCM-Net); (ii)~\texttt{dense-sn}, a dense matrix divided by its largest singular value ($L=1$, but the other singular values can be smaller than one); and (iii)~\texttt{dense}, an unconstrained dense matrix (no bound).
As a reference for accuracy only, we also train an ordinary convolutional network with no Lipschitz constraint, which cannot provide a certificate.

\subsubsection{Orthogonal Adaptation of Pretrained Networks}
\label{subsubsec:pretrained_networks}

In the second setting, a certified network is already available, and the goal is to adapt it to a new task without losing its certificate.
Figure~\ref{fig:method}a,b shows the two certified networks that we adapt and which of their layers are trained.\\

\paragraph{The adapter.}
Let $W$ be a frozen weight matrix of a pretrained network.
The butterfly adapter replaces it by
\begin{equation}
    W' = R(\theta)\,W,
    \label{eq:adapter}
\end{equation}
where $R(\theta)$ is a butterfly circuit acting on the output channels of the layer.
Since $R$ is orthogonal,
\begin{equation}
    W'^\top W' = W^\top R^\top R\, W = W^\top W,
    \label{eq:spectrum}
\end{equation}
so $W'$ has exactly the singular values of $W$, for any $W$ and any angles.
Applied to every adapted layer, this keeps the Lipschitz bound of the whole network at its pretrained value, $L=L_0$, and the certificate stays valid with the same constant.
The frozen part of the network and the frozen weights $W_i$ are shared by all adaptation methods; the butterfly adapter changes only the rotations $R(\theta_i)$ placed in front of them (Fig.~\ref{fig:method}a,b).
The angles start at zero, so $R=I$ and training starts exactly from the pretrained network.
Only the angles and a new classification head are trained.\\

\paragraph{Baselines.}
We compare the butterfly adapter with three other ways of adapting the same layers.
With every method a new classification head is trained, and training starts from exactly the pretrained network.
\begin{itemize}
    \item \textbf{Head-only:} all pretrained weights stay frozen; only the new classifier is trained.
    The bound stays at $L_0$, but the pretrained features are used as they are, and only the classifier on top of them is learned.
    \item \textbf{LoRA:} the frozen weight receives a low-rank update, $W'=W+BA$ with rank $4$ and $B=0$ at the start \cite{hu2022lora}.
    This is parameter-efficient, but nothing controls the singular values of $W+BA$.
    \item \textbf{Full fine-tuning:} the weights of the adapted layers are trained freely.
\end{itemize}

\paragraph{Certified pretrained backbone: BRONet.}
The adapter preserves the bound of whatever network it is applied to, so the starting network has to be certified already.
For the clinical tasks we start from BRONet \cite{lai2025bronet}, a network trained on ImageNet to be certifiably robust; it reaches $49.3\%$ clean and $37.6\%$ certified accuracy at $\varepsilon=36/255$ on ImageNet \cite{lai2025bronet}.
It consists of a convolutional stem, six Lipschitz residual convolution blocks, eight orthogonal block-reflector (BRO) blocks, a neck that pools the feature maps and maps them to $2048$ features with a linear layer, and an eight-layer fully connected network of width $2048$ with MinMax activations, followed by a last-layer-normalised head (see Fig.~\ref{fig:method}b).
The Lipschitz bound of the network up to the fully connected part, measured in the space of $224\times224$ images with pixel values in $[0,1]$, is $L_0=8.11$.
Almost all of it comes from the first convolutional layer (the stem), whose largest singular value is $6.19$; all other layers together contribute a factor of about $1.3$.
We freeze everything up to the fully connected part and adapt its eight $2048\times2048$ layers with the four methods above, together with a new last-layer-normalised head for the target classes.\\

\paragraph{Non-certified control: ResNet152.}
To test whether keeping the bound is enough on its own, we repeat the adaptation on the last bottleneck block of a standard ImageNet-pretrained ResNet152 \cite{he2016deep}, whose Lipschitz bound is of order $10^{109}$ (see App.~\ref{app:resnet}).
This control separates two questions: whether an adapter \emph{keeps} a bound, and whether the bound is small enough to be \emph{useful}.

\subsection{Training Objective}
\label{subsec:loss}

The architecture and the adapter control the denominator of Eq.~\eqref{eq:radius}, the Lipschitz bound.
The numerator, the margin, is determined by training.
A certificate needs a large margin compared with $L$, so the training loss asks for a margin, not only for a correct answer.
In the controlled experiments we use cross-entropy on $8\,(z-\mathbf{1}_y)$: the logit of the correct class $y$ is lowered by one inside the loss, so that the model is rewarded only once its margin exceeds one.
All models built on BRONet, with every adaptation method, are trained with a certificate-aware cross-entropy, a Lipschitz-margin loss \cite{tsuzuku2018lipschitz} with class-pair margins given by the last layer, as in the margin loss of the BRONet code \cite{lai2025bronet}:
\begin{equation}
    \mathcal{L}(x,y)=\mathrm{CE}\big(\tau\,\tilde z(x),\,y\big),
    \label{eq:certce}
\end{equation}
where the logits of the wrong classes are raised before the softmax,
\begin{equation}
    \tilde z_j = z_j + L\,\varepsilon_{\mathrm{train}}\,\|w_y-w_j\|_2
    \quad (j\neq y), \qquad \tilde z_y = z_y ,
    \label{eq:certce_logits}
\end{equation}
and $\tau=4$.
Here $\mathrm{CE}(z,y)=-\log\big(e^{z_y}/\sum_{k=1}^{K}e^{z_k}\big)$ is the cross-entropy between the softmax of the logits $z$ and the true class $y$.
Each wrong class receives the advantage that a perturbation of norm $\varepsilon_{\mathrm{train}}$ could give it, so the loss is small only if the prediction holds within that radius.
We use $\varepsilon_{\mathrm{train}}=36/255$ unless stated otherwise.
The bound $L$ in Eq.~\eqref{eq:certce} is recomputed at the start of every epoch and held fixed during the epoch; it equals $L_0$ for head-only and butterfly adaptation and can grow for LoRA and full fine-tuning.
For ResNet152, the bound is so large that this term would dominate the loss, so the ResNet152 models use the same loss with $\varepsilon_{\mathrm{train}}=0$, i.e.\ cross-entropy on $\tau z$.
The scaling factors in both losses act only inside the loss; they do not change the network or its Lipschitz bound.

% ============================================================
\section{Experimental Setup}
\label{sec:experimental_setup}
% ============================================================

We test the method in two stages.
Controlled experiments on small images show how orthogonality behaves when a network is trained from scratch and when a certified network is adapted.
The trichoscopy experiments then test whether the same mechanism keeps a useful guarantee when a certified network pretrained on ImageNet is adapted to a clinical task. \\

\paragraph{Datasets.}
The controlled experiments use images of $28\times28$ pixels: two binary tasks built from MNIST \cite{lecun1998gradient} (digits 3 versus 5 and 0 versus 8) and two colour datasets from MedMNIST v2 \cite{yang2023medmnist}, BloodMNIST \cite{acevedo2020dataset} and DermaMNIST \cite{tschandl2018ham10000}.
DermaMNIST, with seven classes of skin lesions, is the controlled dataset closest to dermatology; two thirds of its images belong to one class, so we report balanced accuracy on it.
The clinical experiments follow the two trichoscopy tasks of Caposiena Caro \emph{et al.}\ \cite{caro2026trichoscopy}: diagnosis, with five classes---alopecia areata (AA), frontal fibrosing alopecia and lichen planopilaris (FFA-LPP), folliculitis decalvans (FD), androgenetic alopecia (AGA), and a clinically healthy pattern (CHP)---and AA activity, with three classes---active, inactive, and in regrowth.
They contain $3{,}581$ images from 143 patients and $1{,}196$ images from 57 patients.
We use the same preprocessing as the source study \cite{caro2026trichoscopy}: each photograph is cropped to a central $1080\times1080$ square, colour-normalised against a reference image, and resized to $600\times600$ pixels; for the pretrained networks, the images are then resized to $224\times224$, with pixel values in $[0,1]$.
We keep the test sets of the source study unchanged: they contain the original photographs only, 717 images for diagnosis and 240 for AA activity.
The training set of the source study, augmented with rotations, reflections and colour shifts, is split into $90\%$ for training and $10\%$ for validation, with all augmented versions of a in the same part.
The split of the source study is made by image, not by patient; we keep it to stay comparable with that study and discuss its consequences in Sec.~\ref{sec:conclusions}. \\

\paragraph{Training.}
All models are trained using the Adam optimizer \cite{kingma2015adam}. For the networks trained from scratch (Sec.~\ref{subsubsec:orthogonal_networks}), we evaluate all three mixers at depths ranging from 4 to 128 blocks on MNIST 3 versus 5, and from 4 to 64 blocks on DermaMNIST, using two random seeds per configuration. For the controlled transfer experiments, we pretrain a 32-block BCM-Net on MNIST 3 versus 5 and adapt it to MNIST 0 versus 8. We also pretrain a BCM-Net on BloodMNIST and adapt it to DermaMNIST using a class-weighted loss. In both cases, the four fine-tuning strategies are applied to all 32 channel mixers, with two seeds per configuration. We additionally investigate the effect of the margin required by the loss (Sec.~\ref{subsec:loss}) on DermaMNIST transfer, varying it from $0.5$ to $8$.

For the trichoscopy experiments, we apply the fine-tuning strategies to the fully connected layers of BRONet, keeping the feature extractor frozen and computing its outputs only once. Models are trained with a batch size of 256 for up to 30 epochs, using three random seeds per configuration. We select the learning rate for each strategy based on validation accuracy, excluding settings that lead to training collapse (predictions restricted to a single class) or a drop of more than two percentage points in validation accuracy by the final epoch. Among the remaining settings, we select the learning rate and epoch with the highest validation accuracy.

To investigate larger robustness guarantees, we repeat the trichoscopy experiments with $\varepsilon_{\mathrm{train}}\in{72,108,144,195}/255$. In these experiments, the highest validation accuracy often occurs in the first few epochs, before the model has meaningfully adapted. We therefore report the final-epoch model instead.

Finally, we compare classification accuracy against the non-certified convolutional networks from the original trichoscopy study (ResNet152, DenseNet169, and EfficientNetB0), reproduced using the same data split \cite{caro2026trichoscopy}. These baseline models are trained on the full training set and do not provide robustness certificates. \\

\paragraph{Attacks and image transformations.}
We evaluate robustness against adversarial attacks constrained by an $\ell_2$ perturbation budget, matching the norm used for certification. We use PGD \cite{madry2018towards} in all experiments and, for the BRONet models, also evaluate the stronger targeted APGD attack (APGD-T) \cite{croce2020reliable} and random perturbations of the same norm. For trichoscopy, we evaluate the seed-0 model of each fine-tuning strategy on a fixed subset of 300 diagnosis images and all 240 AA-activity test images. For each correctly classified image, we estimate the smallest successful adversarial perturbation $\varepsilon^*(x)$ from Eq.~\eqref{eq:sandwich} by progressively increasing and refining the attack budget. We also verify that no attack succeeds within the certified radius.

Beyond adversarial attacks, we examine common image transformations that may occur in clinical practice, such as JPEG compression, resizing, and reduced bit depth. For each transformation $T$, we measure the resulting change $\|T(x)-x\|_2$ in the same input space used for certification. If this change is smaller than the certified radius $r(x)$, the prediction is guaranteed to remain unchanged. For trichoscopy, we evaluate JPEG compression of the original $600\times600$ images at quality levels ranging from $100$ to $80$. The controlled experiments additionally consider reduced bit depth, resizing, and sensor noise.\\

\paragraph{Evaluation.}
We evaluate each model in terms of classification accuracy and certified robustness. We report accuracy for all datasets and balanced accuracy (the mean recall across classes) for DermaMNIST. Our primary measure of robustness is the \emph{certified accuracy} at radius $\varepsilon$, defined as
\begin{equation}
    \mathrm{CA}(\varepsilon)=\frac{1}{N}\sum_{i=1}^{N}
    \mathbf{1}\!\left[\hat y_i=y_i\;\land\;r(x_i)>\varepsilon\right],
    \label{eq:ca}
\end{equation}
This measures the fraction of test images that are correctly classified and whose predictions are guaranteed to remain unchanged under any perturbation with $\ell_2$ norm up to $\varepsilon$. For trichoscopy, we report certified accuracy at $\varepsilon=36/255$, $72/255$, and $108/255$. For context, $36/255\approx0.14$, whereas changing every pixel of a $224\times224$ colour image by one grey level produces an $\ell_2$ perturbation of approximately $1.5$.

To further characterise robustness, we report the Lipschitz bound after adaptation relative to the pretrained model ($L/L_0$), the median certified radius, and accuracy under adversarial attacks. We also measure the ratio $\varepsilon^*(x)/r(x)$ between the smallest perturbation found by an attack and the certified radius. For image transformations, we report the fraction of correctly classified images whose certificates cover the transformation. Finally, we compare the number of trainable parameters and the measured training time for each strategy.

Results are reported as mean $\pm$ standard deviation across random seeds. To assess statistical significance, we use the exact McNemar test \cite{mcnemar1947note,dietterich1998approximate} to compare strategies on the same test images, combining predictions across seeds by majority vote. We apply the Holm correction \cite{holm1979simple} to account for multiple comparisons between the butterfly adapter and each alternative strategy.
% ============================================================
\section{Results}
\label{sec:results}
% ============================================================

The results are organised as a sequence of questions about the guarantee.
We first ask whether a BCMNet keeps a useful certificate when a network is made deeper (Sec.~\ref{subsec:results_scratch}), and then whether the certificate of a pretrained network survives its adaptation to a new task (Sec.~\ref{subsec:results_finetuning}).
We then move to the clinical setting, the adaptation of a certified ImageNet network to trichoscopy (Sec.~\ref{subsec:results_trichoscopy}), and finally compare the certificate with what attacks actually achieve (Sec.~\ref{subsec:results_attacks}).

\subsection{Networks Trained from Scratch: the Certificate at Increasing Depth}
\label{subsec:results_scratch}

\begin{figure*}[t]
    \centering
    \includegraphics[width=\textwidth]{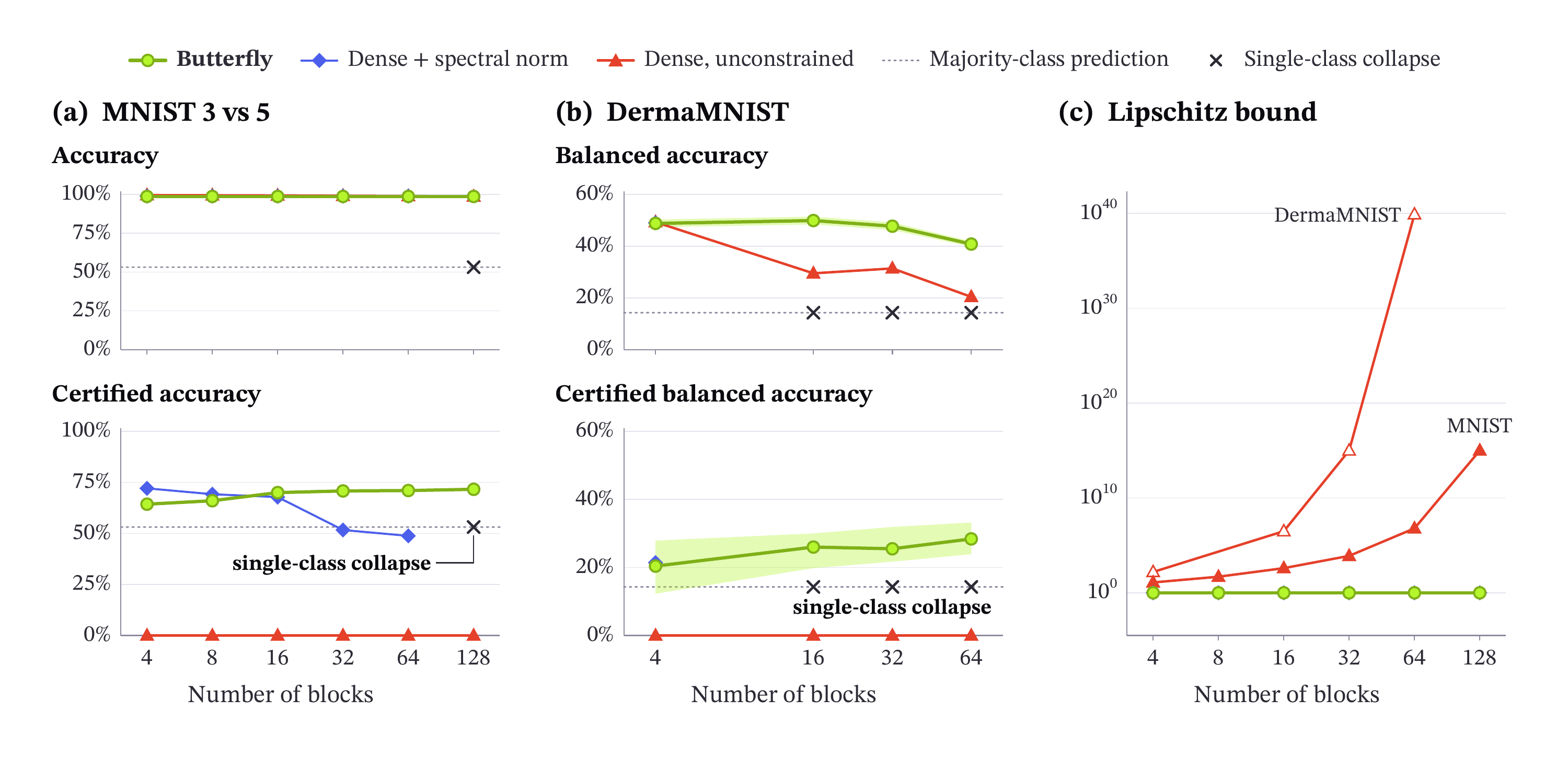}
    \caption{Effect of network depth on classification performance and certified robustness. (a,b) Accuracy (top) and certified accuracy (bottom) for MNIST 3 vs 5 ($\varepsilon=1$) and DermaMNIST ($\varepsilon=0.5$), using balanced metrics for DermaMNIST. Crosses indicate single-class collapse of spectrally normalised networks, and dotted lines show the majority-class baseline. (c) Network Lipschitz bounds, with filled and outlined triangles representing MNIST and DermaMNIST dense networks. }
    \label{fig:controlled}
\end{figure*}

Our first experiments investigate whether deeper networks can maintain useful certified robustness. This requires both a small Lipschitz bound and sufficiently large classification margins: controlling the bound alone is not enough if the network loses its ability to distinguish between classes. We compare BCM-Net, which uses orthogonal butterfly mixers, with spectrally normalised and unconstrained dense mixers at increasing depths on MNIST and DermaMNIST (Fig.~\ref{fig:controlled}). The two conventional approaches illustrate different limitations. Unconstrained dense layers allow the Lipschitz bound to grow rapidly with depth, resulting in negligible certified accuracy at the evaluated perturbation radii. Spectral normalisation avoids this problem by maintaining $L=1$, but classification performance deteriorates as the network becomes deeper, eventually collapsing to predictions of a single class. This is because spectral normalisation limits the expansion of the signal but still allows it to contract, potentially weakening the information needed for classification as layers are stacked. In contrast, orthogonal butterfly mixers preserve the norm of their inputs, allowing the network to maintain a Lipschitz bound of one without the same deterioration in classification performance. On MNIST, accuracy remains high and nearly constant up to 128 blocks, while certified accuracy improves slightly with depth. On DermaMNIST, balanced accuracy decreases at the largest depth, but the network maintains a meaningful certified accuracy throughout. These results suggest that norm-preserving transformations provide a more effective way to maintain useful robustness guarantees in deeper networks than simply constraining their Lipschitz bound.

Having established these properties in networks trained from scratch, we next investigate their potential for fine-tuning. In particular, we examine whether orthogonal butterfly transformations can preserve the Lipschitz bounds of pretrained certified networks while allowing them to adapt to new tasks.

\subsection{Preservation of the Certificate under Adaptation}
\label{subsec:results_finetuning}

We now investigate whether butterfly adapters can preserve certified robustness during fine-tuning. Starting from the 32-block BCM-Nets trained in the previous section, both with $L_0=1$, we adapt them to two new classification tasks using four fine-tuning strategies.

\begin{figure*}[t]
    \centering
    \includegraphics[width=\textwidth]{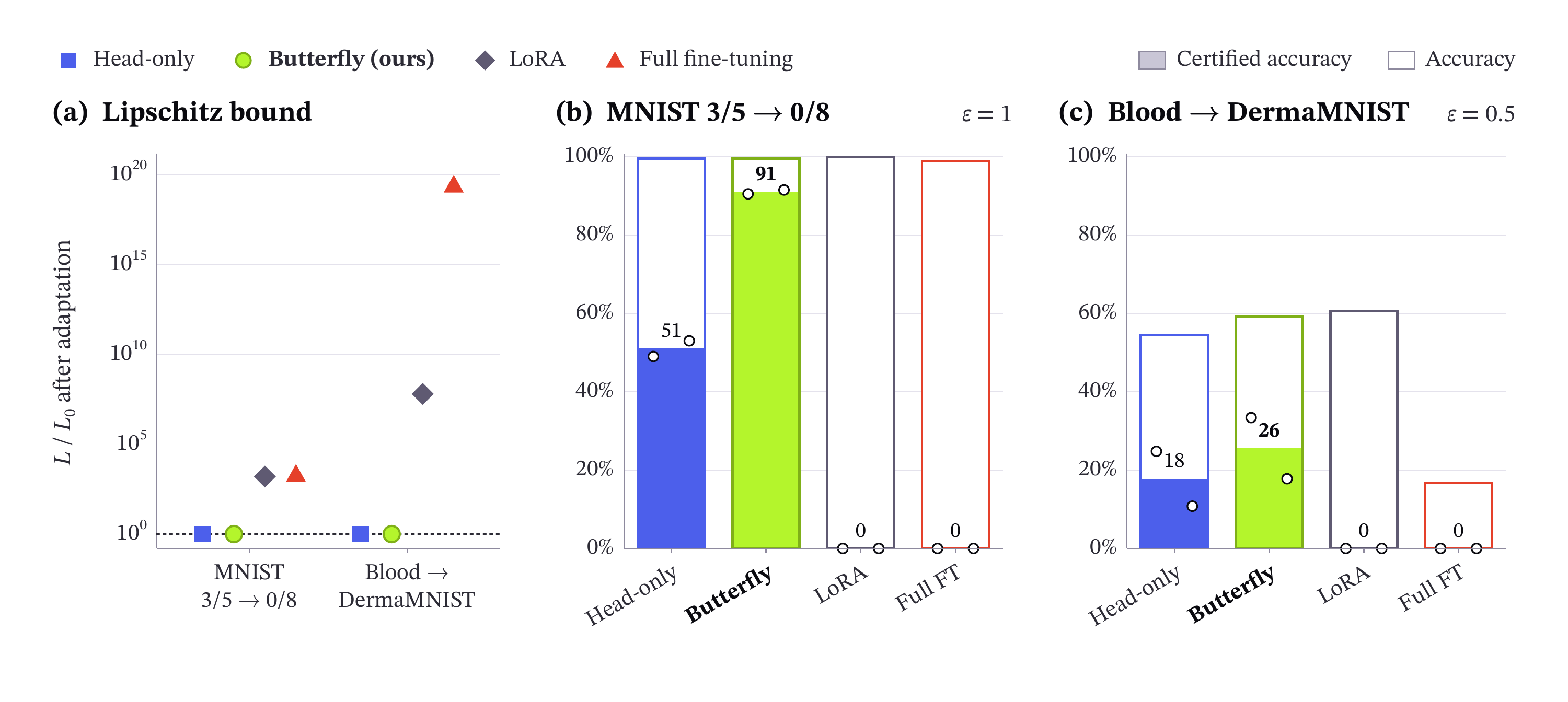}
    \caption{Certified robustness after fine-tuning pretrained BCM-Nets.
(a)~Lipschitz bound relative to the pretrained network, with one marker per seed.
(b,c)~Certified accuracy (filled bars) and accuracy (outlined bars) for MNIST and DermaMNIST transfer at $\varepsilon=1$ and $\varepsilon=0.5$, respectively. Numbers indicate mean certified accuracy, and dots show individual seeds. }
    \label{fig:controlled_transfer}
\end{figure*}

\begin{table*}[t]
\caption{Controlled transfer from a certified 32-block BCM-Net ($L_0=1$). Mean $\pm$ SD over 2 seeds,
last epoch. Accuracy and robust accuracy in \%; median certified radius in input $\ell_2$ units.
Bold: largest median certified radius.}
\label{tab:controlled}
\centering\small
\begin{tabular}{lccccr}
\toprule
Method & Accuracy & $L/L_0$ & Median radius & Accuracy under PGD & Trainable parameters \\
\midrule
\multicolumn{6}{l}{\textit{MNIST 3/5 $\rightarrow$ 0/8} (acc.; PGD at $\varepsilon=1$)} \\
Head-only & 99.1 $\pm$ 0.1 & 1.00 & 1.116 $\pm$ 0.033 & 63.4 $\pm$ 3.1 & 2,050 \\
Butterfly (ours) & 99.4 $\pm$ 0.2 & 1.00 & \textbf{1.760 $\pm$ 0.007} & 92.2 $\pm$ 0.3 & 8,194 \\
LoRA (rank 4) & 99.2 $\pm$ 0.1 & 1529 & 0.003 $\pm$ 0.000 & 47.4 $\pm$ 1.4 & 18,434 \\
Full fine-tuning & 99.1 $\pm$ 0.1 & 1825 & 0.004 $\pm$ 0.001 & 63.0 $\pm$ 5.4 & 135,170 \\
\addlinespace
\multicolumn{6}{l}{\textit{BloodMNIST $\rightarrow$ DermaMNIST} (bal. acc.; PGD at $\varepsilon=1$)} \\
Head-only & 50.6 $\pm$ 4.8 & 1.00 & 0.181 $\pm$ 0.052 & 5.3 $\pm$ 6.1 & 28,679 \\
Butterfly (ours) & 51.5 $\pm$ 2.8 & 1.00 & \textbf{0.214 $\pm$ 0.045} & 11.2 $\pm$ 10.1 & 61,447 \\
LoRA (rank 4) & 54.8 $\pm$ 0.7 & $3.8\times10^{7}$ & 0.000 $\pm$ 0.000 & 0.2 $\pm$ 0.2 & 94,215 \\
Full fine-tuning & 26.5 $\pm$ 6.6 & $1.4\times10^{19}$ & 0.000 $\pm$ 0.000 & 0.0 $\pm$ 0.0 & 2,134,023 \\
\bottomrule
\end{tabular}
\end{table*}

As predicted by the analysis in Sec.~\ref{subsubsec:pretrained_networks} , butterfly adaptation preserves the network's Lipschitz bound. Figure~\ref{fig:controlled_transfer}a  confirms this across all runs: head-only and butterfly adaptation maintain $L/L_0=1$, whereas LoRA and full fine-tuning increase the bound by several orders of magnitude. Crucially, preserving this bound translates into stronger robustness guarantees. On MNIST 3/5 to 0/8, butterfly adaptation achieves 91\% certified accuracy, compared with 51\% for head-only adaptation and nearly zero for LoRA and full fine-tuning, despite similar classification accuracy (Fig.~\ref{fig:controlled_transfer}b and Table~\ref{tab:controlled}). The same pattern appears when transferring from BloodMNIST to DermaMNIST, where butterfly adaptation achieves 26\% certified accuracy compared with 18\% for head-only and zero for the other strategies (Fig.~\ref{fig:controlled_transfer}c). LoRA achieves slightly higher classification accuracy on DermaMNIST, but without comparable guarantees. Butterfly adapters also require fewer trainable parameters than LoRA. We further compare the certified radii with the smallest successful perturbations found by adversarial attacks (Eq.~\eqref{eq:sandwich}). For butterfly and head-only adaptation, these values are relatively close, suggesting that their certificates provide meaningful estimates of robustness. For LoRA and full fine-tuning, the gap spans several orders of magnitude, indicating that their certificates are much more conservative.

\begin{figure}[t]
    \centering
    \includegraphics[width=\columnwidth]{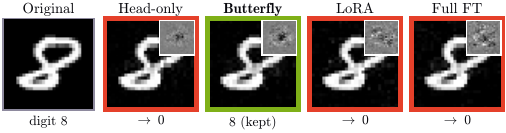}
    \caption{One MNIST 0 vs 8 test image under a worst-case perturbation of the same $\ell_2$ norm, $1.3$ ($13\%$ of the norm of the image), found by PGD separately for each adapted model.
    Insets show the perturbation, amplified to be visible.
    Red frame: the prediction changes; green frame: it is kept.
}
    \label{fig:example}
\end{figure}

Figure~\ref{fig:example} shows what this means for a single image.
Each adapted model receives its own worst-case perturbation of the same small norm: the digit remains clearly recognisable, yet the head-only, LoRA and fully fine-tuned models change their answer, whereas the butterfly model keeps it.

The key advantage of butterfly adaptation is therefore not necessarily greater empirical resistance to adversarial attacks, but the ability to \emph{guarantee} prediction stability. Even when models perform similarly under tested attacks, small, potentially imperceptible perturbations may still change their predictions. Unlike empirical testing, certification guarantees that no perturbation within the certified radius can do so. Having established this advantage in controlled experiments, we next investigate whether it extends to ImageNet-pretrained certified networks adapted to clinical images.

\subsection{Certified Classification of Trichoscopic Images}
\label{subsec:results_trichoscopy}

\begin{figure*}[t]
    \centering
    \includegraphics[width=\textwidth]{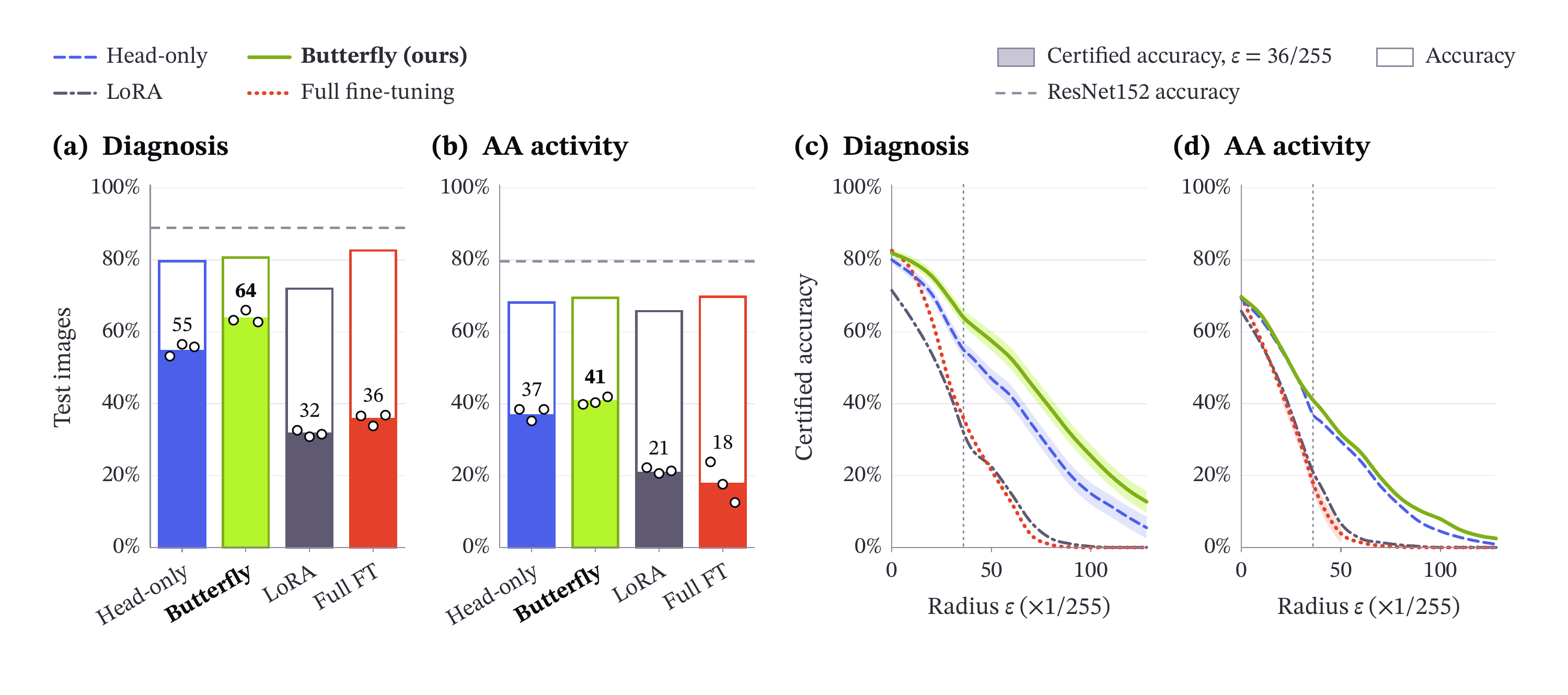}
    \caption{Certified accuracy and classification accuracy on trichoscopy using the certified BRONet backbone.
    (a,b)~Certified accuracy at $\varepsilon=36/255$ (filled bars; numbers indicate the mean, and the three dots indicate different seeds) and classification accuracy (outlined bars) for the four adaptation strategies. The dashed line shows the accuracy of ResNet152 fully fine-tuned as in the source study \cite{caro2026trichoscopy}, which provides no robustness guarantee.
    (c,d)~Certified accuracy as a function of the radius $\varepsilon$; the dotted line marks $36/255$.}
    \label{fig:trichoscopy}
\end{figure*}

We now evaluate butterfly fine-tuning in a clinical setting by adapting the fully connected layers of an ImageNet-pretrained BRONet to two trichoscopy tasks. A useful clinical classifier must first achieve good accuracy, but ideally its predictions should also come with mathematical guarantees of stability. We therefore compare the four adaptation strategies in terms of both accuracy and certified robustness (Fig.~\ref{fig:trichoscopy}). \\

\begin{table*}[t]
\caption{Trichoscopy with the certified BRONet backbone ($L_0=8.11$): accuracy and certified accuracy
(CA) at three radii, in \%. Mean $\pm$ SD over 3 seeds, best validation epoch. $L/L_0$: Lipschitz bound
after adaptation relative to the pretrained network. Time: training steps only, without evaluation and checkpoint writes, one model, on precomputed
features (RTX 4070). $^{\dagger}$Stable learning rate (see Sec.~\ref{sec:experimental_setup}).
Bold: largest certified accuracy at $\varepsilon=36/255$.}
\label{tab:trichoscopy}
\centering\small
\begin{tabular}{lccccccrc}
\toprule
Method & Accuracy & CA@36/255 & CA@72/255 & CA@108/255 & $L/L_0$ & Median radius & Trainable & Time, min \\
\midrule
\multicolumn{9}{l}{\textit{Diagnosis (5 classes, 717 test images)}} \\
Head-only & 79.2 $\pm$ 1.9 & 55.2 $\pm$ 1.8 & 31.3 $\pm$ 4.4 & 11.7 $\pm$ 2.9 & 1.00 $\pm$ 0.00 & 0.193 $\pm$ 0.015 & 10,245 & 0.2 $\pm$ 0.0 \\
Butterfly (ours) & 80.4 $\pm$ 0.4 & \textbf{63.8 $\pm$ 1.7} & 41.5 $\pm$ 4.0 & 20.4 $\pm$ 4.2 & 1.00 $\pm$ 0.00 & 0.251 $\pm$ 0.021 & 100,357 & 14.6 $\pm$ 0.0 \\
LoRA (rank 4)$^{\dagger}$ & 71.5 $\pm$ 0.1 & 31.6 $\pm$ 1.0 & 4.5 $\pm$ 0.3 & 0.0 $\pm$ 0.1 & 1.89 $\pm$ 0.10 & 0.093 $\pm$ 0.005 & 141,317 & 0.7 $\pm$ 0.0 \\
Full fine-tuning$^{\dagger}$ & 82.5 $\pm$ 0.6 & 35.6 $\pm$ 1.7 & 1.7 $\pm$ 0.2 & 0.0 $\pm$ 0.0 & 5.12 $\pm$ 0.77 & 0.111 $\pm$ 0.002 & 33,581,061 & 1.1 $\pm$ 0.0 \\
\addlinespace
\multicolumn{9}{l}{\textit{AA activity (3 classes, 240 test images)}} \\
Head-only & 67.8 $\pm$ 0.9 & 37.2 $\pm$ 1.9 & 13.5 $\pm$ 1.5 & 2.1 $\pm$ 0.0 & 1.00 $\pm$ 0.00 & 0.132 $\pm$ 0.007 & 6,147 & 0.1 $\pm$ 0.0 \\
Butterfly (ours) & 69.3 $\pm$ 2.1 & \textbf{40.6 $\pm$ 1.3} & 17.2 $\pm$ 0.6 & 4.7 $\pm$ 0.2 & 1.00 $\pm$ 0.00 & 0.135 $\pm$ 0.006 & 96,259 & 6.4 $\pm$ 0.0 \\
LoRA (rank 4)$^{\dagger}$ & 65.3 $\pm$ 0.2 & 21.3 $\pm$ 0.8 & 1.0 $\pm$ 0.6 & 0.0 $\pm$ 0.0 & 2.47 $\pm$ 0.31 & 0.082 $\pm$ 0.004 & 137,219 & 0.3 $\pm$ 0.0 \\
Full fine-tuning$^{\dagger}$ & 69.6 $\pm$ 0.8 & 17.9 $\pm$ 5.6 & 0.7 $\pm$ 0.5 & 0.0 $\pm$ 0.0 & 8.49 $\pm$ 5.05 & 0.074 $\pm$ 0.005 & 33,576,963 & 0.5 $\pm$ 0.0 \\
\bottomrule
\end{tabular}
\end{table*}

\paragraph{Accuracy.}
All four strategies achieve comparable classification accuracy (Fig.~\ref{fig:trichoscopy}a,b). For diagnosis, butterfly adaptation significantly outperforms LoRA and performs slightly worse than full fine-tuning. For AA activity, the differences are not statistically significant. Based on accuracy alone, there is therefore little reason to prefer one adaptation strategy over another. \\

\paragraph{Certified accuracy.}
The differences become much more pronounced when considering certified accuracy. For diagnosis, butterfly adaptation correctly classifies and certifies approximately two thirds of the test images at $\varepsilon=36/255$, compared with only one third for LoRA and full fine-tuning. A similar twofold improvement is observed for AA activity. These differences are highly significant on both tasks and become even more pronounced at larger radii (Fig.~\ref{fig:trichoscopy}c,d). At $\varepsilon=108/255$, butterfly adaptation still certifies approximately 20\% of diagnosis images, while LoRA and full fine-tuning certify none.

Butterfly adaptation also outperforms head-only training, despite both strategies preserving the Lipschitz bound. This improvement is significant for diagnosis, although not for AA activity with its 240 test images. These results show that orthogonal butterfly adapters can improve certified accuracy by adapting the learned representations without increasing the network's Lipschitz bound. \\

\paragraph{Why the other strategies lose the guarantee.}
As in the controlled experiments, the main difference lies in the Lipschitz bound. LoRA and full fine-tuning increase it by a factor of two to eight (Table~\ref{tab:trichoscopy}). Since the certified radius depends on both the classification margin and the Lipschitz bound (Eq.~\eqref{eq:radius}), this increase weakens the guarantees unless compensated by larger margins. In our experiments, the margins do not compensate sufficiently, leaving LoRA and full fine-tuning with median certified radii roughly half those achieved by butterfly adaptation. \\

\paragraph{Parameters and training time.}
Butterfly adaptation achieves this higher certified accuracy with approximately $10^5$ trainable parameters, fewer than LoRA and over 300 times fewer than full fine-tuning. This may be useful in medical applications, where labelled datasets are often small and fewer trainable parameters could limit overfitting, although we did not evaluate this effect here. However, this parameter efficiency does not translate into faster training. In our implementation, each training step constructs eight $2048\times2048$ orthogonal matrices from their rotation angles, making butterfly adaptation the slowest strategy (Table~\ref{tab:trichoscopy}). Future works includes developing more time-efficient implementation of quantum-inspired orthogonal adaptations. \\

\paragraph{Accuracy limitations of the pretrained backbone.}
On the same image-level split, the certified models are less accurate than the non-certified convolutional networks of the source study (Fig.~\ref{fig:trichoscopy}a,b); this comparison concerns different backbones and training conditions and should be interpreted with care. An important limitation is the pretrained backbone: BRONet achieves $49.3\%$ top-1 accuracy on ImageNet \cite{lai2025bronet}, substantially below standard architectures such as ResNet152 \cite{he2016deep}. The comparison also differs in training data, since the source models use the full training set, whereas ours use $90\%$. These results highlight the current limitations in classification accuracy of our certified models, which future improvements in certified backbones may help overcome. \\

\paragraph{Size of the guarantee.}
Beyond certified accuracy at a fixed radius, we investigate whether the guarantees can be made large enough to cover common image transformations. In the controlled DermaMNIST transfer experiment, increasing the margin required by the loss doubles the median certified radius of the butterfly model without reducing balanced accuracy. Its default certificates already cover JPEG re-encoding, reduced bit depth, moderate resizing, and sensor noise for most test images, while LoRA and full fine-tuning cover none (App.~\ref{app:controlled_transforms}, Fig.~\ref{fig:controlled_transforms}). \\

\begin{figure*}[t]
    \centering
    \includegraphics[width=\textwidth]{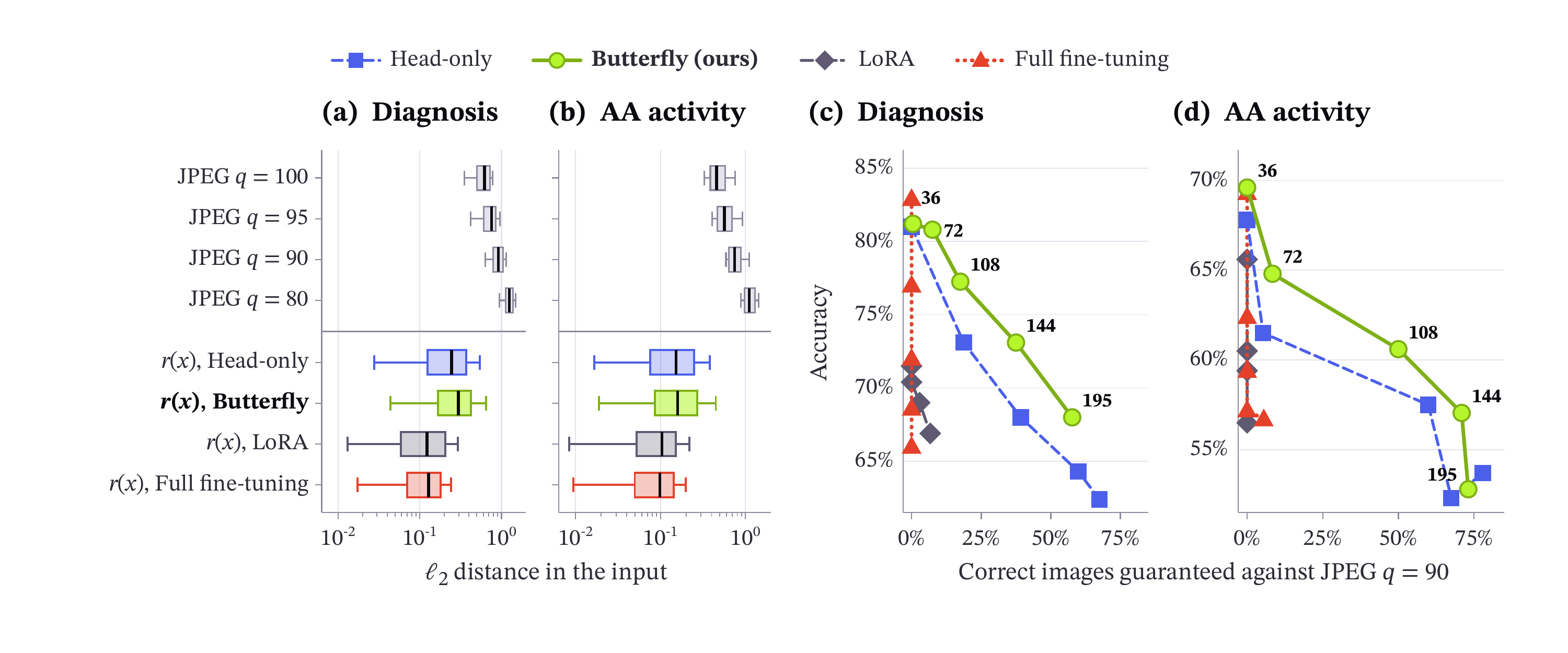}
    \caption{JPEG re-encoding and the size of the guarantee on trichoscopy.
    (a,b)~$\ell_2$ change of the input caused by re-encoding the image file as JPEG (grey) and certified radius of correctly classified images (coloured), for $\varepsilon_{\mathrm{train}}=36/255$; boxes show the quartiles, whiskers the 5th and 95th percentiles.
    (c,d)~Accuracy against the share of correctly classified images whose certificate covers JPEG quality 90, for $\varepsilon_{\mathrm{train}}\in\{36,72,108,144,195\}/255$ (numbers on the butterfly curve); for AA activity, LoRA has no point at $195/255$ because all three seeds collapsed.
    BRONet, models \emph{after the last training epoch} (with a large $\varepsilon_{\mathrm{train}}$ the best validation epoch is the first or second one); three seeds, seeds that collapsed to a single class excluded.}
    \label{fig:jpeg}
\end{figure*}

For trichoscopy, however, the default certificates are too small to cover typical image-processing operations. JPEG re-encoding at quality 90 changes the network input by several times the typical certified radius (Fig.~\ref{fig:jpeg}a,b). Consequently, only a small fraction of images are certified against this transformation, even though JPEG compression rarely changes the predicted class. This is partly due to image resolution: the $\ell_2$ norm accumulates small changes across all $224\times224\times3$ input values.

To obtain larger guarantees, we train the models with increasing target radii $\varepsilon_{\mathrm{train}}$. For butterfly and head-only adaptation, the certified radii increase almost proportionally. At the largest target radius, butterfly certificates cover JPEG quality 90 for most correctly classified images, although classification accuracy decreases by 13 to 17 percentage points (Fig.~\ref{fig:jpeg}c,d). In contrast, LoRA and full fine-tuning achieve less than $7\%$ coverage across all target radii, as the growth of their Lipschitz bounds offsets the larger classification margins. Thus, stronger guarantees are possible, but on trichoscopy they come at a substantial cost in accuracy.\\

\paragraph{Control with a non-certified backbone.}
Finally, we investigate whether butterfly adaptation can provide useful guarantees when applied to a conventional, non-certified network such as ResNet152. Although the adapter preserves the backbone's Lipschitz bound, this bound is approximately $10^{109}$, making the resulting certificates ineffective. Moreover, an adversarial attack with $\varepsilon=9/255$ already reduces classification accuracy by half (App.~\ref{app:resnet}, Table~\ref{tab:resnet}). This confirms an important limitation: orthogonal adapters can preserve the Lipschitz bound of an existing network, but they cannot turn a non-certified backbone into a meaningfully certified one.

\subsection{Certified versus Empirical Robustness}
\label{subsec:results_attacks}

Although LoRA and full fine-tuning provide much weaker robustness guarantees than butterfly adaptation, this does not necessarily mean that their models are more vulnerable to adversarial attacks. To distinguish between empirical robustness and what can be mathematically guaranteed, we evaluate the seed-0 model of each strategy using PGD and APGD-T, comparing the certified radii with the perturbations found by the attacks (App.~\ref{app:attacks}, Fig.~\ref{fig:attacks_curves}). The four strategies show similar empirical robustness. At $\varepsilon=36/255$, a comparable fraction of images withstands PGD across all methods, with APGD-T producing only slightly stronger attacks. Random perturbations of the same norm do not change any predictions. Importantly, no attack succeeds within the certified radius, consistent with the mathematical guarantees. The main difference is therefore not how the models respond to the tested attacks, but how much of their robustness can be certified.

This distinction is also evident at the individual-image level. For butterfly and head-only adaptation, the certified radii are strongly correlated with the smallest successful perturbations found by the attacks (Spearman correlation $0.97$--$0.99$), indicating that larger certificates generally correspond to images that are harder to attack. For LoRA and full fine-tuning, the ratio between the smallest successful perturbation and the certified radius is approximately twice as large, despite similar empirical attack thresholds. Their increased Lipschitz bounds therefore lead to more conservative certificates, without a corresponding increase in observed vulnerability.

On trichoscopy, the advantage of butterfly adaptation is thus not greater empirical resistance to the attacks tested, but a larger share of predictions whose stability is mathematically certified. While empirical attacks cannot rule out smaller, potentially imperceptible perturbations that change a diagnosis, certification guarantees stability against \emph{any} perturbation within the certified radius. This contrasts with the controlled DermaMNIST transfer, where butterfly adaptation also improved empirical robustness. One possible explanation is that those experiments adapted all 32 mixers of BCM-Net, whereas the trichoscopy experiments modify only the final layers of BRONet.

\section{Conclusions}
\label{sec:conclusions}

In this work, we investigated quantum-inspired orthogonal butterfly layers for building and adapting certifiably robust neural networks, with trichoscopy as our main clinical application. In networks trained from scratch, butterfly layers maintain useful certified robustness as depth increases, avoiding the deterioration observed with spectral normalisation and the growing Lipschitz bounds of unconstrained networks. More importantly, we demonstrated that butterfly adapters can fine-tune pretrained certified networks without increasing their Lipschitz bounds. On alopecia diagnosis and alopecia areata activity assessment, this approach achieves approximately twice the certified accuracy of LoRA and full fine-tuning, with comparable classification accuracy among the adaptation methods evaluated on the same backbone and data split. In these experiments, butterfly adaptation achieved higher certified accuracy than LoRA and unrestricted fine-tuning, while the methods showed similar empirical resistance to the tested adversarial attacks.

These guarantees address an important limitation of conventional model evaluation. While test accuracy and adversarial attacks provide empirical evidence of reliability, certification guarantees that no perturbation within a specified radius can change an individual prediction. This does not establish that the diagnosis is medically correct, nor does it cover changes outside the certified region. Nevertheless, it provides a form of verifiable stability that could support robustness assessments of medical AI \cite{euaiact2024,fda2025aienabled}. Certified radii could also help identify predictions that warrant additional clinical review, although the usefulness of such a strategy remains to be evaluated.

An important advantage of our approach is that it combines these guarantees with parameter-efficient adaptation. Butterfly transformations preserve orthogonality by construction using only $\tfrac{n}{2}\log_2 n$ trainable angles, an established property of these quantum-inspired architectures. Their reduced parameter count may also help limit overfitting on small medical datasets. The main drawback in our implementation is training time, although the adapted weights can be computed once after training, allowing inference at the speed of the original network.

The quality of the pretrained backbone remains a major limitation. Our certified models achieve lower accuracy than conventional convolutional networks, and obtaining larger certificates on trichoscopic images introduces a further accuracy trade-off. However, our method is not specific to BRONet: it can be applied to other networks with known, useful Lipschitz bounds \cite{trockman2021orthogonalizing,singla2021skew,hu2023unlocking,lai2025bronet}. This opens an interesting direction for quantum-inspired architectures. Beyond using butterfly circuits as fine-tuning adapters, future work could explore designing and pretraining larger certified networks built from such orthogonal structures. Building on the properties demonstrated by BCM-Net in our controlled experiments, these architectures could provide more accurate pretrained backbones that retain their robustness guarantees and can subsequently be adapted to specialized medical tasks. Their correspondence to shallow quantum circuits also provides a foundation for exploring future quantum implementations, although practical advantages remain to be established.

This study has several limitations. First, we followed the image-level split used in the original study, which allows images from the same patient to appear in both training and test sets, potentially leading to overestimated accuracy \cite{yagis2021effect,tampu2022inflation}. We retained this split to focus on certified robustness and compare adaptation methods under the same experimental conditions as the original study. Consequently, the reported accuracies should not be interpreted as estimates of performance on unseen patients, which requires further validation using patient-level splits and larger datasets. The clinical datasets are also relatively small (143 and 57 patients), and adversarial attacks were evaluated using a single seed per strategy and a subset of test images. Finally, the clinical experiments focused on adapting the fully connected layers of BRONet, without investigating the effects of acquisition variations such as illumination conditions or imaging devices.

Future work should prioritize patient-level validation on larger, independent datasets, more efficient implementations of butterfly adapters, and guarantees covering a wider range of clinically relevant image variations. Improving the accuracy of certified pretrained backbones, potentially through larger quantum-inspired orthogonal networks, will also be essential for making these approaches clinically useful. Overall, our results indicate that quantum-inspired butterfly adapters offer a parameter-efficient way to adapt certified neural networks without compromising their Lipschitz bounds. By preserving mathematical guarantees while allowing models to learn new tasks, they provide a promising step towards medical image classifiers with verifiable prediction stability.

\section*{Code availability}
The code used to train and evaluate the models is publicly available at \url{https://github.com/eigenorbit/quantum-inspired-fine-tuning}.

\bibliography{references}

% ============================================================
\appendix
% ============================================================

% The appendix has more floats than text: pack them at the top of float pages
% (no stretched gaps between them) and allow more of them on one page.
\makeatletter
\setlength{\@fptop}{0pt}
\setlength{\@fpsep}{12pt plus 2pt}
\setlength{\@fpbot}{0pt plus 1fil}
\setlength{\@dblfptop}{0pt}
\setlength{\@dblfpsep}{12pt plus 2pt}
\setlength{\@dblfpbot}{0pt plus 1fil}
\makeatother
\renewcommand{\topfraction}{0.95}
\renewcommand{\dbltopfraction}{0.95}
\renewcommand{\bottomfraction}{0.8}
\renewcommand{\textfraction}{0.05}
\renewcommand{\floatpagefraction}{0.6}
\renewcommand{\dblfloatpagefraction}{0.6}
\setcounter{topnumber}{3}
\setcounter{dbltopnumber}{3}
\setcounter{bottomnumber}{2}
\setcounter{totalnumber}{6}

\section{Control: Adaptation of a Non-Certified Backbone}
\label{app:resnet}

\begin{table}[!htbp]
\caption{Control without a certified backbone: ResNet152 with its last bottleneck block adapted
(batch 32). Accuracy in \%, mean $\pm$ SD over 3 seeds, best validation epoch; $L/L_0$: mean over seeds.
Certified accuracy is $0\%$ for every model at every radius. PGD: accuracy under attack, in \%,
at $\varepsilon=9/255$ and $36/255$ (seed 0). Params: trainable parameters.
$^{\dagger}$Stable learning rate (see Sec.~\ref{sec:experimental_setup}).}
\label{tab:resnet}
\centering\footnotesize
\setlength{\tabcolsep}{3.5pt}
\resizebox{\columnwidth}{!}{%
\begin{tabular}{lccccr}
\toprule
 & & & \multicolumn{2}{c}{PGD} & \\
\cmidrule(lr){4-5}
Method & Accuracy & $L/L_0$ & 9/255 & 36/255 & Params \\
\midrule
\multicolumn{6}{l}{\textit{Diagnosis} ($\log_{10} L_0 = 108.9$)} \\
Head-only & 81.0 $\pm$ 0.7 & 1 & 56.7 & 11.0 & 10k \\
Butterfly & 83.6 $\pm$ 0.5 & 1 & 57.3 & 7.3 & 26k \\
LoRA & 82.8 $\pm$ 0.3 & 12.73 & 51.3 & 6.3 & 51k \\
Full FT & 84.9 $\pm$ 1.2 & 1.02 & 59.3 & 15.0 & 4,473k \\
\addlinespace
\multicolumn{6}{l}{\textit{AA activity} ($\log_{10} L_0 = 108.9$)} \\
Head-only & 76.7 $\pm$ 3.0 & 1 & 48.3 & 5.4 & 6k \\
Butterfly$^{\dagger}$ & 79.0 $\pm$ 1.0 & 1 & 47.9 & 6.2 & 22k \\
LoRA & 76.8 $\pm$ 1.0 & 3.02 & 41.2 & 2.1 & 47k \\
Full FT & 76.4 $\pm$ 1.7 & 3.68 & 46.7 & 7.1 & 4,469k \\
\bottomrule
\end{tabular}}
\end{table}

Since butterfly adaptation preserves the bound of any network, it could in principle be applied to a standard, more accurate network instead of a certified one.
To test whether this gives a useful guarantee, we repeat the adaptation with an ImageNet-pretrained ResNet152 \cite{he2016deep}.
We freeze the network up to its last bottleneck block and adapt the three convolutions of that block (with batch normalisation folded into the weights); the butterfly adapter rotates the output channels of each convolution, and a new last-layer-normalised head is trained in every case.

On ResNet152, butterfly adaptation again keeps the Lipschitz bound exactly, and the models are more accurate than those built on BRONet (Table~\ref{tab:resnet}).
But the bound that is kept is about $10^{109}$, so every certified radius is about $10^{-109}$ and no prediction of any model is guaranteed.
The lack of a guarantee is not only theoretical: an attack with $\varepsilon=9/255$, four times smaller than the radius used for BRONet, already reduces the accuracy of these models to about one half, and at $36/255$ to $2$--$15\%$ (Table~\ref{tab:resnet}), whereas the BRONet models keep $58$--$78\%$ under the same attack.

\section{Proven versus Empirical Robustness on Trichoscopy}
\label{app:attacks}

\begin{figure*}[!t]
    \centering
    \includegraphics[width=\textwidth]{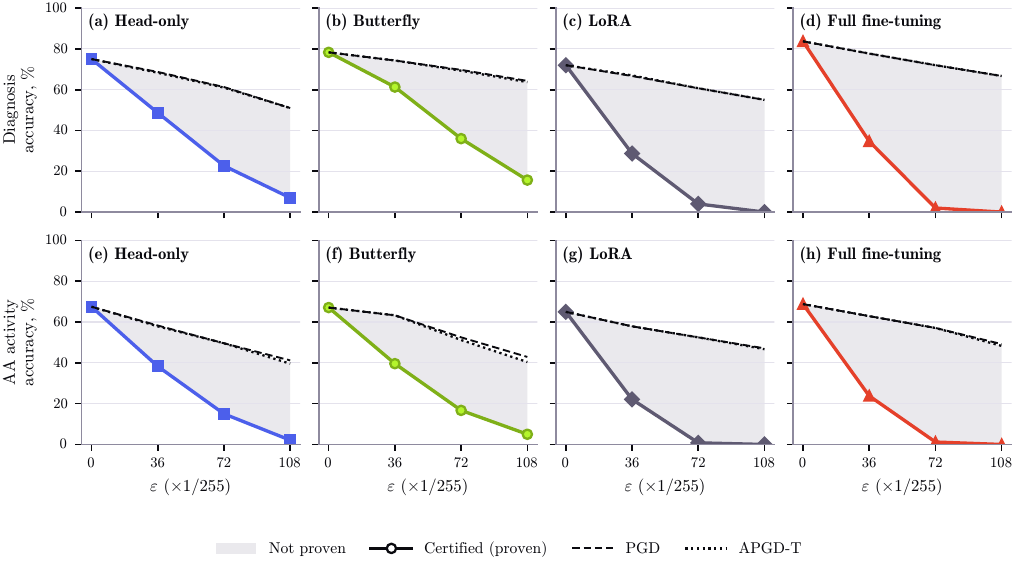}
    \caption{Proven versus empirical robustness on trichoscopy.
    (a--d)~Diagnosis and (e--h)~AA activity, one panel per adaptation strategy.
    Coloured lines: certified accuracy, a lower bound on the true robust accuracy; black lines: accuracy under PGD (dashed) and APGD-T (dotted), upper bounds.
    The grey area between them is robustness that the models show under attack but that is not proven.
    BRONet, seed 0, best validation epoch; 300 diagnosis and 240 AA-activity test images.}
    \label{fig:attacks_curves}
\end{figure*}

Section~\ref{subsec:results_attacks} summarises how the certificate relates to adversarial attacks on trichoscopy; here we give the setting and the numbers behind that summary. \\

\begin{figure*}[!t]
    \centering
    \includegraphics[width=\textwidth]{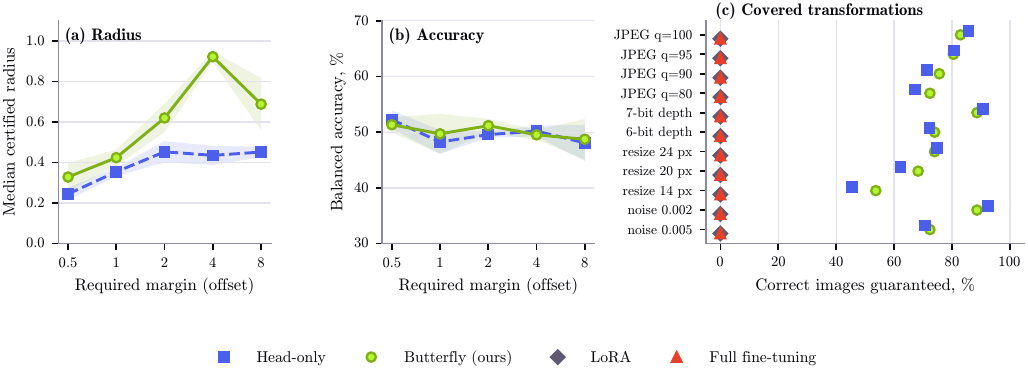}
    \caption{Size of the guarantee in the controlled transfer BloodMNIST $\rightarrow$ DermaMNIST.
    (a)~Median certified radius and (b)~balanced accuracy as a function of the margin required by the loss; $L=1$ throughout.
    (c)~Share of correctly classified test images whose certificate covers each image transformation, with the default margin; LoRA and full fine-tuning cover no image for any transformation.
    Last epoch; (a,b)~three seeds, bands show $\pm$SD; (c)~one model per strategy, 200 test images.}
    \label{fig:controlled_transforms}
\end{figure*}

\paragraph{Setting.}
We attack the seed-0 model of each strategy on a fixed subset of 300 diagnosis and all 240 AA-activity test images, the same images for every model.
We use PGD with 20 steps and one restart \cite{madry2018towards} at $\varepsilon\in\{36,72,108\}/255$, the targeted APGD attack (APGD-T, 100 iterations) \cite{croce2020reliable}, and, as a comparison, random perturbations of exactly the same norm.
All attacks search for an $\ell_2$-bounded perturbation and are not clipped to $[0,1]$, so that they test exactly the set of perturbations covered by the certificate.
To estimate the smallest successful perturbation $\varepsilon^*(x)$ of Eq.~\eqref{eq:sandwich} for a correctly classified image, we start from its certified radius, double the attack budget until the attack succeeds, and refine it by eight bisection steps. \\

\paragraph{The certificate is never violated.}
Over all models, radii and attacks, no successful perturbation was smaller than the certified radius, with either the official or a rigorous upper estimate of $L_0$.
This confirms in practice that the bound and the certificates are computed correctly. \\

\paragraph{Under attack, all strategies behave similarly.}
The black lines in Fig.~\ref{fig:attacks_curves} are close for all four strategies.
At $\varepsilon=36/255$, between $67\%$ and $78\%$ of the diagnosis images and between $58\%$ and $63\%$ of the AA-activity images withstand PGD.
The stronger APGD-T attack lowers these numbers by at most three points, so PGD does not substantially overestimate the robustness, and random noise of the same size changes nothing. \\

\paragraph{The difference lies in what is proven.}
The coloured lines in Fig.~\ref{fig:attacks_curves} differ much more than the black ones.
For butterfly adaptation, the certificate proves most of the robustness that the attacks observe at $\varepsilon=36/255$ (for diagnosis, $61\%$ proven of $74\%$ observed), and the grey area is about as small as for head-only training.
For LoRA and full fine-tuning, most of the observed robustness is not proven (for diagnosis, $29$--$35\%$ proven of $67$--$78\%$ observed), and at $\varepsilon=72/255$ almost none of it is. \\

\paragraph{Single images.}
For butterfly and head-only adaptation, the smallest successful attack is about $3.5$--$3.8$ times the certified radius, and the two quantities are almost perfectly ranked together (Spearman correlation $0.97$--$0.99$).
For LoRA and full fine-tuning, this ratio rises to $7.5$--$9.3$, although the smallest successful attack itself is similar or even somewhat larger.
Their larger Lipschitz bound therefore does not make these models more fragile; it makes their certificate weaker.

\section{Size of the Guarantee in the Controlled Transfer}
\label{app:controlled_transforms}

Figure~\ref{fig:controlled_transforms} gives the results behind the paragraph on the size of the guarantee in Sec.~\ref{subsec:results_trichoscopy}.\\

\paragraph{Larger margin.}
We repeat the BloodMNIST $\rightarrow$ DermaMNIST transfer for butterfly and head-only adaptation with the margin required by the loss (Sec.~\ref{subsec:loss}) set to $0.5$, $1$, $2$, $4$ and $8$, with three seeds.
For the butterfly model, a margin of $4$ instead of the default $1$ doubles the median certified radius, from $0.42$ to $0.92$, without any loss of balanced accuracy (Fig.~\ref{fig:controlled_transforms}a,b).
The radius of the head-only model, in contrast, stops growing at about $0.45$: only the butterfly adapter can rotate the features to allow larger margins.\\

\paragraph{Image transformations.}
On 200 test images, we apply JPEG re-encoding with quality $100$ to $80$, 7- and 6-bit quantisation, resizing to $24$, $20$ and $14$ pixels and back, and Gaussian sensor noise with standard deviation $0.002$ and $0.005$, and check whether the change of the input is smaller than the certified radius.
At this low resolution, already with the default margin, the certificate of the butterfly model covers JPEG down to quality 80, reduced bit depth, moderate resizing and sensor noise for $68$--$89\%$ of the correctly classified images, while LoRA and full fine-tuning cover none (Fig.~\ref{fig:controlled_transforms}c).
The model that proves nothing is also the one that changes its predictions most often under these operations (up to $10.5\%$ for full fine-tuning, at most $1.5\%$ for the butterfly model).

\end{document}